\documentclass[letterpaper]{article} 
\usepackage{aaai2026}  
\usepackage{times}  
\usepackage{helvet}  
\usepackage{courier}  
\usepackage[hyphens]{url}  
\usepackage{graphicx} 
\usepackage{natbib}  
\usepackage{caption} 
\usepackage{algorithm}
\usepackage{algorithmic}
\usepackage{amsmath}
\usepackage{amsfonts}
\usepackage{bm}
\usepackage{amsthm}
\usepackage{booktabs}
\usepackage{newfloat}
\usepackage{listings}
\DeclareCaptionStyle{ruled}{labelfont=normalfont,labelsep=colon,strut=off} 
\floatstyle{ruled}
\newfloat{listing}{tb}{lst}{}
\floatname{listing}{Listing}
\title{ERPPO: Entropy Regularization-based Proximal Policy Optimization}
\author{
    Changha Lee\textsuperscript{\rm 1},
    Gyusang Cho\textsuperscript{\rm 1}
}
\affiliations{
    \textsuperscript{\rm 1}Korea Advanced Institute of Science and Technology, Daejeon, South Korea\\

}

\usepackage{bibentry}

\begin{document}

\maketitle

\begin{abstract}
Multi-Agent Proximal Policy Optimization (MAPPO) is a variant of the Proximal Policy Optimization (PPO) algorithm, specifically tailored for multi-agent reinforcement learning (MARL). MAPPO optimizes cooperative multi-agent settings by employing a centralized critic with decentralized actors. However, in case of multi-dimensional environment, MAPPO can not extract  optimal policy due to non-stationary agent observation. To overcome this problem, we introduce a novel approach, Entropy Regularization-based Proximal Policy Optimization (ERPPO). For the policy optimization, we first define the object detection ambiguity under multi-dimensional observation environment. Distributional Spatiotemporal Ambiguity (DSA) learner is trained to estimate object detection uncertainty in non-stationary constraints. Then, we enhance PPO with a novel Entropy Regularization term. This regularization dynamically adjusts the policy update by applying a stronger (L1) regularization in high-ambiguity observation to encourage significant exploratory actions and a weaker (L2) regularization in low-ambiguity observation to stabilize the proximal policy optimization. This approach is designed to enhance the probability of successful object localization in time-critical operations by reducing detection failures and optimizing search policy. Experiments on a testbed with AirSim-based maritime searching scenarios show that the proposed ERPPO improves accuracy performance. Our proposed method improves higher gradient than MAPPO. Qualitative results confirm that ERPPO effectiveness in terms of suppressing false detection in visually uncertain conditions.
\end{abstract}


\section{Introduction}


Proximal Policy Optimization (PPO) is a widely adopted on‑policy policy‑gradient method noted for its balance between empirical performance and implementation simplicity~\cite{schulman2017proximal}. Building on PPO, the Multi‑Agent Proximal Policy Optimization (MAPPO) framework extends these benefits to cooperative multi‑agent reinforcement learning (MARL) tasks and has achieved state‑of‑the‑art results on benchmarks~\cite{yu2022surprising}. MAPPO follows the centralized training with decentralized execution paradigm, coupling a single centralized critic—conditioned on the joint observation with multiple decentralized actors that operate on local observations at test time~\cite{amato2024introduction}. This design improves sample efficiency and mitigates the credit‑assignment problem by leveraging global information during training~\cite{kapoor2024assigning}, yet retains scalability because each agent’s policy remains lightweight and fully distributed during execution.

For the adapting dynamic environment in Multi-UAV search and controlling~\cite{abdelnabi2024human,gromada2022real}, MAPPO-based approaches have been researched. Despite its effectiveness in cooperative multi-agent scenarios, MAPPO encounters significant limitations when applied to multi-dimensional environments characterized by non-stationary observations. In such complex settings, each agent's observation space dynamically changes, influenced by the simultaneous actions and interactions of other agents, thus creating a non-stationary context~\cite{papoudakis2019dealing,nekoei2023dealing}. Consequently, the centralized critic in MAPPO struggles to accurately assess joint state-action values, leading to suboptimal policy extraction. This non-stationarity disrupts the stability and convergence properties typically guaranteed by MAPPO, emphasizing the need for further research and methodological advancements to enhance its robustness in highly dynamic, multi-dimensional MARL environments.

To address these challenges, we propose an entropy regularization-based proximal policy optimization (ERPPO) with spatiotemporal ambiguity measurement. To adapt dynamic environment, we dynamically retrains RL policies governing UAV operations, enabling rapid adaptation to evolving mission scenarios. UAVs, guided by these updated policies, execute real-time, latency-critical surveillance tasks by deep learning-based object detection. In particular, to support AI-specific sophisticated training techniques, we have focused on studies related to maritime search and rescue (MSAR) which is one of time-critical applications.

The key contributions of this work are two-fold:
\begin{itemize}
 \item \textbf{Distributional Ambiguity Learning:} We propose distributional spatiotemporal ambiguity (DSA) learner by computing the confidence field in the simulation environment, to infer the ambiguity in dynamic environment.
 
 \item \textbf{Entropy regularization-based Proximal Policy Optimization:}  We propose entropy regularization in policy optimization. Proposed method regularizes the policy loss with entropy estimation, thereby guarantee object localization under environmental uncertainty.
\end{itemize}

\section{Related Works and Problem Description}
In this section, we introduce the related works and describe an ambiguity problem.

\subsection{Object Detection for Untrained Environment}

For the better part of a decade, deep learning has revolutionized object detection, producing models with remarkable accuracy in identifying and localizing objects within images. The success of architectures like YOLO~\cite{redmon2016you}, SSD~\cite{liu2016ssd}, and Faster R-CNN~\cite{ren2015faster} has been largely demonstrated in closed-set scenarios, where models are trained and tested on data drawn from the same, well-defined distribution with a fixed set of object categories. However, the real world is not a closed set; it is a dynamic, unpredictable, and ever-changing open environment.

\subsubsection{Environment-Specialized Object Detection:}
To handle the dynamic environment in object detection model, there are several researches. YoloOW~\cite{xu2024yoloow} enhance the detection of missing ships or persons, where the orientation of the target is critical for accurate localization and identification in vast maritime environments. They solve problem: (i) variations in the appearance and size of objects (people, vessels) depending on altitude and viewing angle. (ii) false detections caused by reflections on the water surface. These method successfully achieved higher probability in the maritime object detection for UAV aerial imagery.

\subsubsection{Open-Vocabulary Object Detection:}
Open-vocabulary object detection represents a significant paradigm shift in computer vision, moving beyond the limitations of traditional closed-set detection systems. It aims to detect objects from an arbitrary, user-defined set of categories at inference time, without being explicitly trained on them. open-vocabulary object detection~\cite{wang2023detecting, cheng2024yolo} largely attributed to advancements in vision-language pre-training, particularly with models like CLIP (Contrastive Language-Image Pre-training)~\cite{radford2021learning}. These models learn a shared embedding space where images and their corresponding text descriptions are aligned. The fundamental principle of most open-vocabulary object detection is to transfer this learned vision-language knowledge to a dedicated object detector architecture.

\begin{figure*}[!h]
    \centering
    \includegraphics[width=1\linewidth]{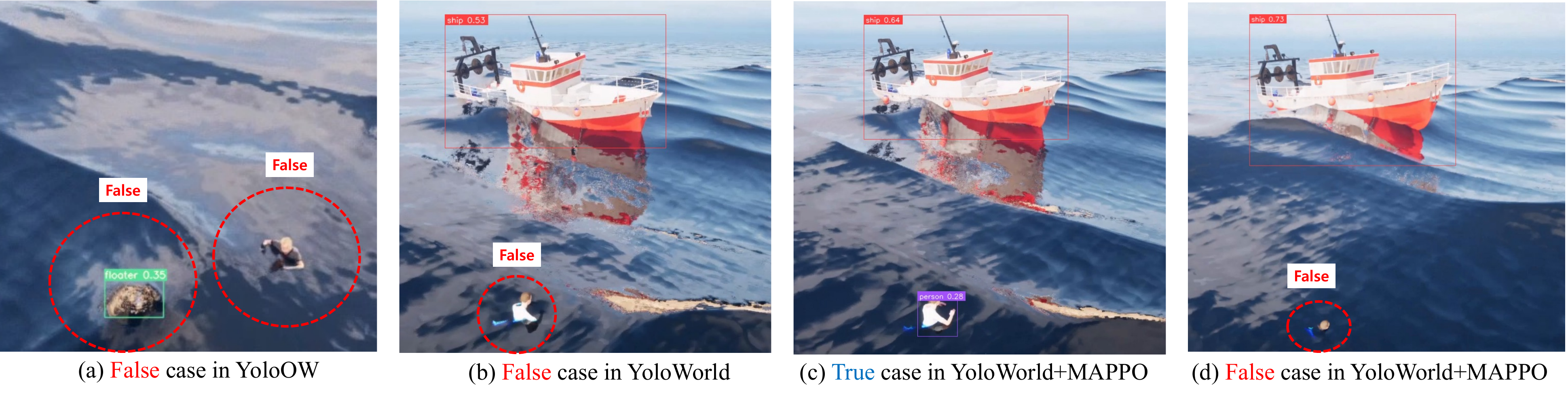}
    \vspace{-6mm}
    \caption{False object detection results generated by (a) YoloOW~\cite{xu2024yoloow} and (b) YoloWorld~\cite{cheng2024yolo}. (c) shows the true case and movement by MAPPO~\cite{yu2022surprising}. Even though applying MAPPO, (d) represents the false case.}
    \label{fig:prob1}
\end{figure*}

\subsection{Problem Description}
Despite the development of sophisticated object detection models~\cite{xu2024yoloow,cheng2024yolo,zhao2023revisiting,zohar2023prob}, encountering failure cases in real-world applications is inevitable. It is unrealistic to expect any AI model to achieve 100\% accuracy, and object detection models are no exception. As illustrated in Figure~\ref{fig:prob1}, these failure cases persist even with existing research focused on developing environment-specific object detection models~\cite{xu2024yoloow,cheng2024yolo} or those capable of detecting objects in an open-vocabulary setting.

Previous researches have largely focused on evaluating and improving object detection performance in broader open-world scenarios. However, we proposes an alternative approach, getting inspiration from human perception. We posit that interacting with the real world, much like humans adjust their viewpoint to resolve ambiguities, can effectively reduce object detection failure cases.

This leads to our formalized problem definition: If the camera's pose can be represented as a state, then actions such as changing the camera's position or reducing its distance to an object can be viewed as deliberate strategies to mitigate ambiguity in object detection.

\subsection{Spatiotemporal Ambiguity in Object Detection}

Let $o_k$ be the target object position and $t$ be the time. The locational probability of the target object $o_k$ at time $t_k$ is defined as $p(o_k,t)$. In this work, we measure the locational probability of object in the image obtained by a camera. Let the image be $\mathbf{Y}(x_u,t)$ where $x_u$ is UAV position. Then, the observed locational probability is formulated as follows:
\begin{equation}
    p(o_k,x_u,t)= p\Big(o_k=f_\theta(\mathbf{Y}(x_u,t))\Big)
    \label{eq:prob}
\end{equation}
where $f_{\theta}$ represents object detection inference result. Equation~\eqref{eq:prob} means the probability that object detection inference result equals the target object $o_k$. 

From this locational probability, we applied the Shannon Entropy to quantify the ambiguity. Shannon entropy, as known formula is $H(o,x,t)=-p(o,x,t)\log p(o,x,t)$, converges to 0 for probabilities that can be determined with certainty, while to 1 for probabilities that are \textit{ambiguous}. 

Here, we define the ambiguous masking array for an object $o$ as \textit{ambiguous} when the obtained maximum value does not exceed a certain threshold $\sigma$ as follows: 

\begin{equation}
\label{eq:amb}
    \mathcal{M}(x,t)= \begin{cases} \textit{ambiguous}, \quad \qquad\underset{o\in \mathbb{R}}{\text{max }}{H(o,x,t)} \geq \sigma, \\
    \textit{non-ambiguous},\quad \underset{o\in \mathbb{R}}{\text{max }}{H(o,x,t)} < \sigma,
    \end{cases}
\end{equation}
where the actual value of \textit{ambiguous} is 1 and \textit{non-ambiguous} is 0.
The main purpose of the proposed framework is to reduce ambiguity and enhance searching performance. Reducing ambiguity is important as it greatly reduces failures.  To modeling locational ambiguity, we assume that the maritime dynamic environment is mainly affected by four factors: (1) weather, (2) the location of the UAV, (3) the movement of objects by waves, and (4) the reflection of the wave surface. Accordingly, we attempted to solve the ambiguity due to time and space based on the formalization of the weather and UAV observation space change on the wave simulator.

\subsection{Unobserved Locational Ambiguity: Not Decreasing}
As time progresses in the simulation, if an object is continuously influenced by its maritime environment, where waves affect its position, the resulting change in the object's location can be described as a diffusion process over time as follows:
\begin{equation}
    p(o_k,x,t)=\prod_{i=0}^{t-1}p(o_k,x,i+1|i)
\end{equation}

As the probability of a specific location outcome satisfies $0 \leq p(o,x,i+1|i)\leq 1$ for all $i$, we can conclude that:
\begin{equation}
    p(o_k,x,t)\leq p(o_k,x,t+1), \quad \forall t
    \label{eq:ineqa}
\end{equation}
The proof of inequality in Equation~\eqref{eq:ineqa} is in Supplementary. This implies that location ambiguity is increasing or equals. From this condition, our model successfully reduces the ambiguity in the policy and value model optimization.

\section{Proposed Method}
In this section, we consider a decentralized partially observable Markov decision process (Dec-POMDP)~\cite{oliehoek2016concise}. Therefore, we introduce the partially observable non-stationary environment causing false object detection. we discuss the proposed DSA learner model's training process in terms of ambiguity issue with simulated observations. With regard of reinforcement learning-based optimization issue through DSA analysis, we suggested the entropy regularization for searching task.
\begin{figure*}
    \centering
    \includegraphics[width=1\linewidth]{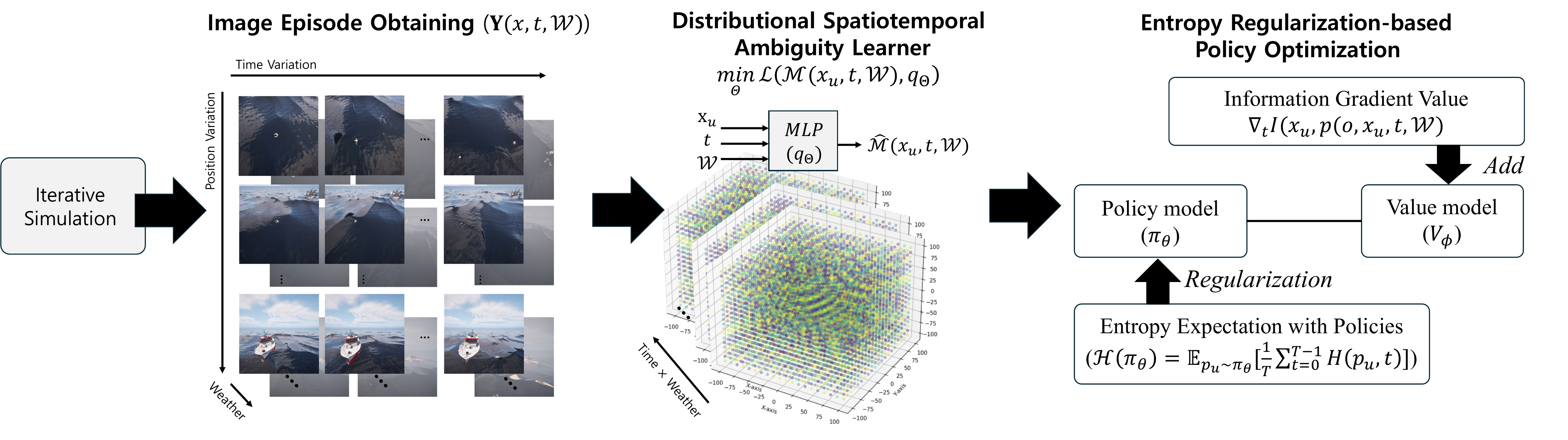}
    \caption{Overview of the proposed framework with distributional spatiotemporal ambiguity (DSA) learner and entropy regularization-based proximal policy optimization (ERPPO).}
    \label{fig:arch}
\end{figure*}

\subsection{Partially Observable Dynamic Environment} 
At time step $t$, UAV $i$ observes its state $s_i$ and select an action $a_i \in \mathcal{A}_i$. After that, UAV $i$ generates its own trajectory $\tau_i$, which could be shared to agents around it. The joint actions of agents $a_t = (a^1_t, \cdots, a^N_t)$ yield the new environment state $s'$ and immediate reward $\{r^i_t\}_{i=1}^N$ according to the transition probability $\mathcal{T}: \mathcal{S} \times \mathcal{A} \times \mathcal{S} \rightarrow [0,1]$ and reward function $\mathcal{S} \times \mathcal{A} \rightarrow \mathbb{R}$, respectively, in which $\mathcal{S}$ and $\mathcal{A}$ are the state space and action space of system (i.e. $\mathcal{S} = \prod_{i=1}^M \mathcal{S}_i$ and $\mathcal{A} = \prod_{i=1}^M \mathcal{A}_i$). The main goal of each agent is to learn an optimal policy $\pi^*$ which can maximize the expected return.

In Dec-POMDP, each UAVs collect the locational probability score according to an UAV's path. Let UAV's camera image obtained at the $i$-th UAV's position $x_i$ at time $t$ be $\mathbf{Y}_t =  \{\mathbf{Y}(x_1,t), \cdots, \mathbf{Y}(x_M,t)\}$. The object location information of the image at time $t_j$ is given as  $\mathbf{o}_{j}^i = (u_{j}^i, v_{j}^i, w_{j}^i, h_{j}^i)$, and have classification result $\hat{c}(\mathbf{o}_{j}^i,t_j)$. $(u_j^i, v_j^i)$ is the coordinate of the center of bounding box, $(w_{j}^i, h_{j}^i)$ means the width and height of the bounding box. It has a locational probability score at each UAV position $p(\mathbf{o}_{j}^i,t_i)$. In simulator environment, each weathers (e.g. rain or fog) can be represented by the value $\mathcal{W}\in[0,1]$. For example, $\mathcal{W}_{\rm{rain}}=0$ means no rain, and $\mathcal{W}_{\rm{fog}}=1$ is heavy fog that is maximum value in simulation. 

\subsection{Distributional Spatiotemporal Ambiguity Learner}
To predict the spatiotemporal ambiguity in not only simulation but also open-world, we propose the construction of a Distributional Spatiotemporal Ambiguity (DSA) learner.

The learning process begins with a preliminary data acquisition. A UAV is deployed to perform a systematic survey of the search area, capturing a comprehensive set of images $\mathbf{Y}$ at given each positions $p_u$, time $t$, and weather values $\mathcal{W}$. These captured images are then processed using a powerful open-vocabulary object detection model. In our framework, we utilize object detection model with YoloWorld~\cite{cheng2024yolo}, which can identify objects beyond its pre-defined training categories by leveraging textual prompts (e.g., "a person in water"). For each detected object $o_k$, the model provides a bounding box and a confidence score. 

From obtained all images, we need to know the distribution of object locational ambiguity, however the real distribution is hard to know because all observations are partial and discrete observation. Therefore, we propose the spatiotemporal ambiguity predictor with the surrogate $q$-distribution learner $q_\Theta$ as follows:
\begin{equation}
    \hat{\mathcal{M}}=q_\Theta(x,t,\mathcal{W})
\end{equation}
where $\hat{\mathcal{M}}$ represents the estimated ambiguity by the DSA Leaner. To train the trainable parameter $\Theta$, the loss function for DSA learner is formulated by:
\begin{equation}
L_\Theta=||\mathcal{M}(x,t,\mathcal{W}),q_\Theta(x_u,t,\mathcal{W})||_2+\frac{\lambda}{2} |\Theta|^2
\end{equation}
where $q_\Theta$ represents a multi-layer perceptron with trainable parameter $\Theta$. Here, ambiguities $\mathcal{M}(x,t,\mathcal{W})$ are aggregated across all collected images to form the DSA. In the next section, the DSA learner predicts the ambiguity map $\hat{\mathcal{M}}(x,t,\mathcal{W})$.  This pre-computed map provides the RL agent with a strong inductive bias, enabling it to formulate a more intelligent and efficient search strategy from the very first step of its mission.

\subsection{Proximal Policy Optimization with Entropy Regularization}
The estimated result by DSA learner is negligible given that the probabilities derived here are based on simulations and may exhibit a completely different distribution from the actual search scenario. Nevertheless, this paper assumes that the direction of the search can still be known when determining the policy of UAVs based on the search expectations calculated from the DSA learner.

\subsubsection{(1) Information Gradient Value Estimation:}
Multiple UAVs searches the region of interest in maritime environment. To maximize the searching probability, we can setup training objective in RL with the object locational probability $p(o,x,t,\mathcal{W})$. Let the $I\big(x_i,t,p(o,x_i,t,\mathcal{W})\big)$ be the obtained information of $i$-th UAV search when the UAV position is $x_i$ at time $t$ and $p(x_i,t,\mathcal{W})$ is the observed locational probability in Equation~\eqref{eq:prob}. To approximate $I\big(x_i,t,p(o,x_i,t,\mathcal{W})\big)$, we consider the searching area with the weather composite $\bar{W}= \mathcal{W}_{rain} + \mathcal{W}_{fog}$ where $r_{rain}$ is the raindrop ratio and $r_{fog}$ is fog rate by referred to ~\cite{gultepe2010visibility}. The approximated information quantity of all searching object $o$ is given by:
\begin{equation}
\begin{split}
    &I \big(x_u,p(o,x_u,t,\bar{W})\big) \\ 
    &= \sum_{o_k\in \mathbf{O}} \sum_{x_u} \frac{p\Big(o_k=f_\theta(\mathbf{Y}(x_u,t,\bar{W}))\Big)}{||p\Big(o_k=f_\theta(\mathbf{Y}(x_u,t,\bar{W}))\Big)||^2}
\end{split}
\end{equation}

Now, the value model should predict the value according to the state transition. Therefore, the target value is suggested as follows:
\begin{equation}
\begin{split}
V_\phi (t)= I(x_u,p(x_u,t+1,\bar{W}))-I(x_u,p(x_u,t,\bar{W}))
\end{split}
\label{eq:value}
\end{equation}
Equation~\eqref{eq:value} means that target value is valuable when there is the positive difference of information according to a policy. Detailed reward and value function are explained in Supplementary.

\begin{figure*}
    \centering
    \includegraphics[width=\linewidth]{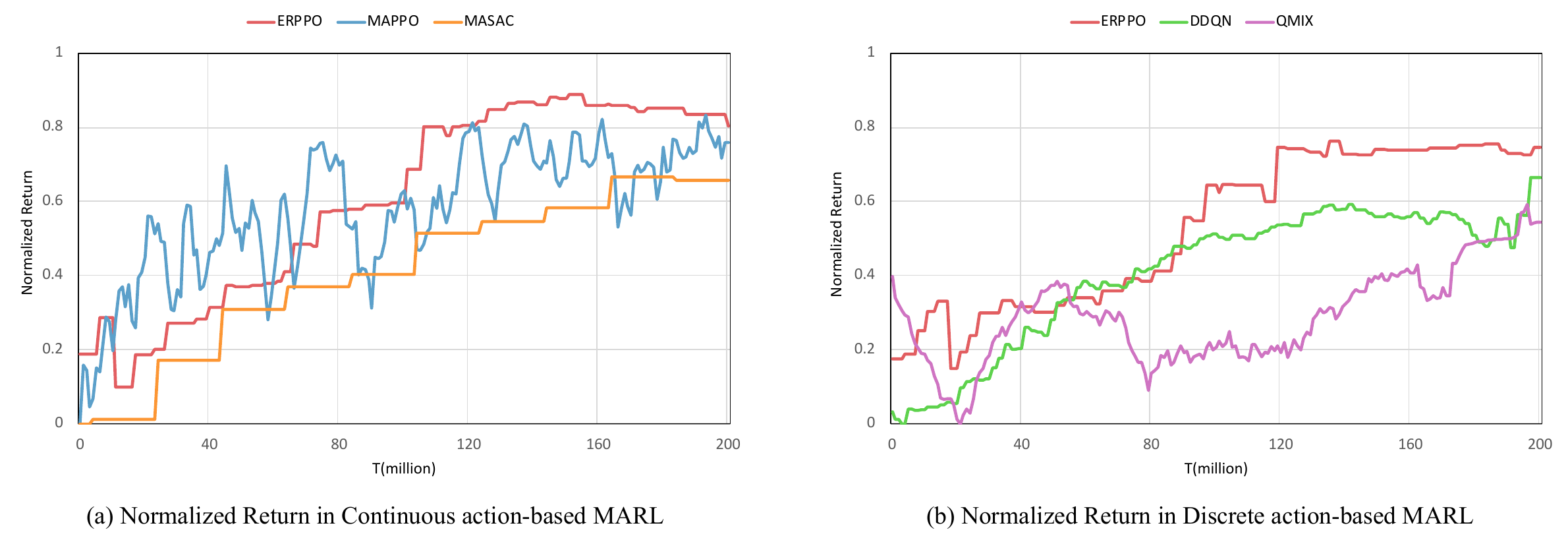}
    \caption{Performance comparison of ERPPO with MAPPO, MASAC, DDQN, and QMIX in partially observable maritime dynamic environment}
    \label{fig:graph1}
\end{figure*}

\subsubsection{(2) Entropy Gradient Policy Regularization:}
We propose the policy optimization scheme considering the entropy. To update the policy network, we concern entropy expectation at first. According to the policy model $\pi_\theta$, the entire expectation of entropy $\mathcal{H}$ is defined as follows:
\begin{equation}
    \mathcal{H}(\pi_\theta)=\mathbb{E}_{x_u\sim \pi_\theta}[ \mathbb{E}_t[H(o,x_u,t)]]
    \label{eq:exp_entropy}
\end{equation}

Using Equation~\eqref{eq:exp_entropy} with estimated ambiguity $\hat{\mathcal{M}}$, the new approach ERPPO is formulated using the following equation:
\begin{equation}
\small
    \begin{split}
        L(\theta )= \begin{cases}
            L^{\rm{CLIP}}+|\mathcal{H}(\pi_\theta)-\mathcal{H}(\pi_{\theta_{k}})|,\ \text{if}\ \mathbb{E}[\hat{\mathcal{M}}] \geq \sigma.\\
            L^{\rm{CLIP}}+\frac{1}{2}|\mathcal{H}(\pi_\theta)-\mathcal{H}(\pi_{\theta_{k}})|^2,\ \text{otherwise.}
        \end{cases}
    \end{split}
    \label{eq:mappo2}
\end{equation}

In other words, higher entropy values induce more important updates from previous policies to new ones by using L1 regularization, while lower entropy values facilitate the retention of previous policies with L2 regularization. 

Algorithm~\ref{alg:samegr} details our proposed method, Multi-agent ERPPO, which integrates a novel regularization term into the MAPPO~\cite{yu2022surprising}. To adapt entropy regularization, our training scheme starts from the initial policy and value model parameters, $\theta_0$ and $\phi_0$, respectively. After all training iterations, we finally obtain the optimal value function $V^*_\phi$ and policy model $\pi_\theta^*$.

\begin{algorithm}[!ht]
\caption{ERPPO Algorithm}
\label{alg:samegr}
\begin{algorithmic}[1]
\STATE \textbf{Initialize:} value $V_{\phi_0}$, policy $\pi_{\theta_0}$.
\FOR{each training iteration}
    \FOR{each UAV $j$ at each time step $t$}
        \STATE Estimate Spatiotemporal Ambiguity Map $\hat{\mathcal{M}}$.
        \STATE Calculate search information $I(x_u,p(x_u,t,\bar{W}))$.
        \STATE Store rollouts in a replay buffer.
    \ENDFOR
    \STATE Update value function $V_\phi$ and policy model $\pi_\theta$.
    \STATE $\text{minimize}_\phi\ L(\phi)=\mathbb{E}_t[V_\phi(t)-R_t]$ (Refer to Supplementary)
    \STATE $\text{minimize}_\theta\ L(\theta)$\\
    $=\begin{cases}
         L^{\rm{CLIP}}+|\mathcal{H}(\pi_\theta)-\mathcal{H}(\pi_{\theta_{k}})|,\quad\text{if}\ \mathbb{E}[\mathcal{M}] \geq \sigma.\\
            L^{\rm{CLIP}}+\frac{1}{2}|\mathcal{H}(\pi_\theta)-\mathcal{H}(\pi_{\theta_{k}})|^2,\ \text{otherwise.}
    \end{cases}$
\ENDFOR
\STATE \textbf{return} Optimized value function $V_{\phi}^*$ policy model $\pi_{\theta}^*$
\end{algorithmic}
\end{algorithm}

    



\section{Experiments}
This section details the environment, hardware configuration, training parameters, performance metrics, and evaluation methodology employed in our study.

\subsection{RL and DL Environment}
Experiments were conducted using the multi-UAVs search and rescue environment, a classic control problem, sourced from the Gymnasium reinforcement learning library. We implemented the MAPPO algorithm, leveraging Ray RLlib~\cite{liang2018rllib} framework for distributed training. For visualization and simulation of environment, we exploited AirSim~\cite{airsim2017fsr} with Unreal Engine~\cite{unrealengine}. For the YOLO-based experiment, we implemented the pretrained YOLOv8-world~\cite{cheng2024yolo}, using Pytorch~\cite{paszke2019pytorch}.

\subsection{Hardware Configuration}
The experiments were executed on a system equipped with Dual Intel(R) Xeon(R) Silver 4110 CPUs, each operating at a base frequency of 2.10 GHz. Each CPU features 8 physical cores and 16 threads, contributing to a total of 16 physical cores and 32 logical processors ($N_{cpu}$) available to the system. The system's memory (RAM) is approximately 128 GB. For graphical processing, a total of four GPUs were available ($M_{gpu}$), which included two NVIDIA GeForce RTX 2080 GPUs, each with 8 GB of VRAM, and two NVIDIA GeForce RTX 2080 Ti GPUs, each with 11 GB of VRAM. The installed NVIDIA driver version was 530.30.02, supporting CUDA version 12.1. Furthermore, the system was configured to support a maximum of 8 parallel Unmanned Aerial Vehicle (UAV) simulation workers ($M_{uav}$).


\subsection{Performance comparison of ERPPO and other MARL}
In this section, we detail the evaluation of our proposed method alongside several baseline Multi-Agent Reinforcement Learning (MARL) algorithms. To this end, we establish a simulated maritime environment, which is characterized as a partially observable and dynamic system. We benchmark the performance of our approach against prominent MARL techniques, which are categorized based on whether they operate in a continuous or discrete action space.

\subsubsection{Baselines: } The selected baseline methods are divided into two main categories as follows:

\begin{itemize}
    \item \textbf{Continuous action space:} (1) Multi-Agent PPO (MAPPO)~\cite{yu2022surprising} extends this framework to multi-agent settings, where each agent independently learns its policy via PPO, often sharing a centralized value function for more stable learning. (2) Multi-Agent Soft Actor-Critic (MASAC)~\cite{lowe2017multi} is an off-policy algorithm grounded in the maximum entropy reinforcement learning framework. It aims to maximize both the expected reward and the policy's entropy, which encourages robust exploration and leads to more stable and sample-efficient learning, particularly in complex continuous control tasks.
    \item \textbf{Discrete action space:} (1) Multi-agent Double Deep Q-Network (DADDQN)~\cite{van2016deep} is applied using an Independent Q-Learning (IQL) approach, where each agent maintains its own DDQN to learn its optimal action-value function. (2) QMIX~\cite{rashid2020monotonic} utilizes a mixing network that estimates the joint action-value function as a monotonic combination of each agent's individual Q-values, ensuring consistent credit assignment during decentralized execution.
\end{itemize}

\begin{figure*}[t!]
    \centering
    \includegraphics[width=\linewidth]{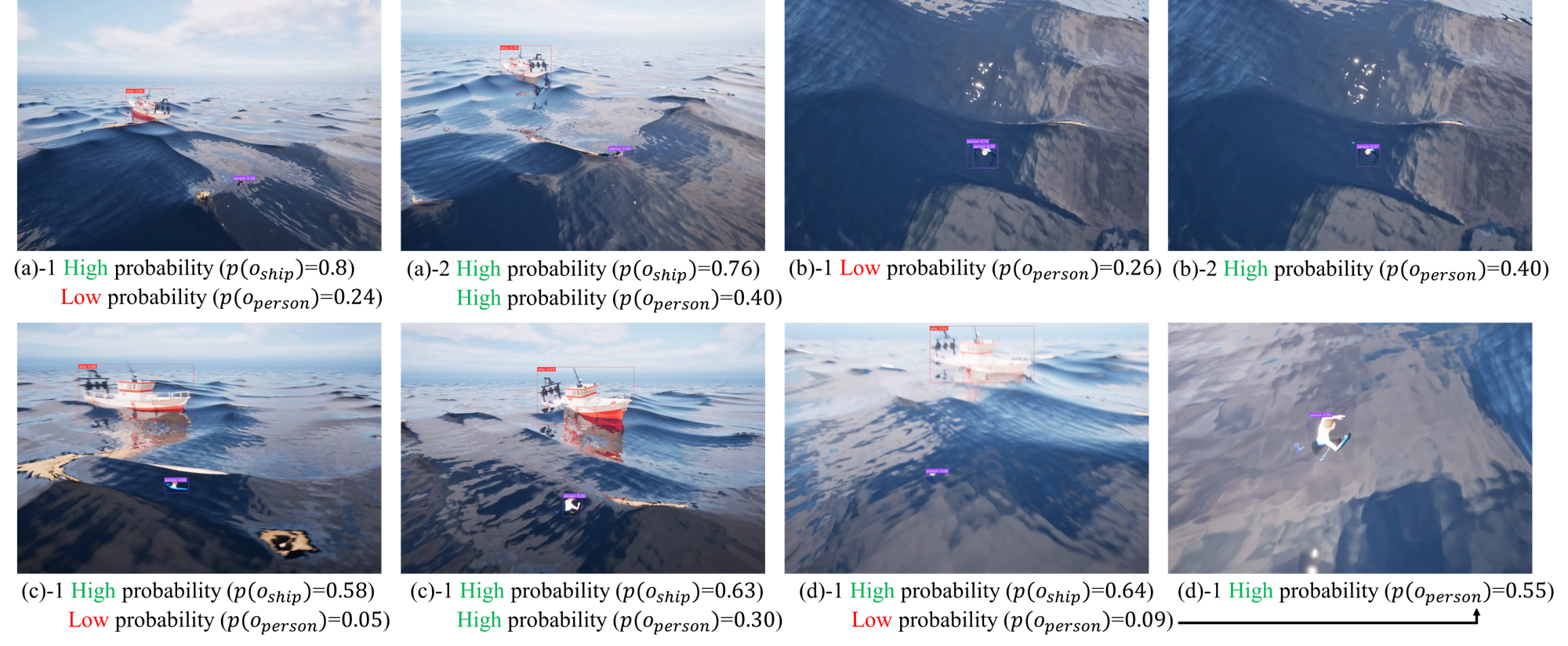}
    \caption{Qualitative results of the ambiguity mitigation by UAVs control in maritime scenarios using AirSim~\cite{airsim2017fsr} with Unreal Engine~\cite{unrealengine}. Green text indicates 'High object locational probability' detections, while red text indicates 'Low object locational probability' detections, accompanied by localization scores.}
    \label{fig:qualitative_ambiguity_results}
\end{figure*}

\subsubsection{Configurations: } All algorithms, including our proposed method, were trained for a total of 200 million timesteps to ensure a fair comparison. Key hyperparameters, such as the learning rate, were primarily set according to the values recommended in their respective reference papers~\cite{yu2022surprising,lowe2017multi,van2016deep,rashid2020monotonic}. A comprehensive list and detailed explanation of all hyperparameter configurations are provided in the Supplementary.

\subsubsection{Discussion: } Figure~\ref{fig:graph1} presents the comparative performance of our proposed ERPPO algorithm against several baseline MARL methods, evaluated over 200 million training timesteps. The performance is quantified by the normalized return, where a higher value indicates better performance.

Figure~\ref{fig:graph1}(a) illustrates the learning curves for algorithms operating in a continuous action space: ERPPO, MAPPO, and MASAC. Our proposed ERPPO method demonstrates a robust and stable learning trajectory, achieving the highest normalized return which consistently stays above 0.8 after 120 million timesteps. In contrast, while MAPPO exhibits rapid initial learning, its performance is marked by high variance and significant oscillations throughout the training process. MASAC shows a more stable learning progression than MAPPO but converges to a suboptimal normalized return of approximately 0.65. This result highlights ERPPO's superior stability and convergence performance in complex continuous control environments.

Figure~\ref{fig:graph1}(b) displays the results for the discrete action space, comparing ERPPO against DDQN and QMIX. Similar to the continuous setting, ERPPO achieves the most favorable outcome, reaching a stable normalized return of approximately 0.75 by the 120 million timestep mark. DDQN demonstrates steady learning, gradually improving its performance over time and eventually surpassing QMIX, reaching a return of nearly 0.7 by the end of the training. Conversely, QMIX suffers from significant learning instability; after an initial phase, its performance degrades and remains erratic, ultimately converging to the lowest return among the discrete-action methods.

\subsection{Ablation study of Object Locational Probability Change in ERPPO}

Figure~\ref{fig:qualitative_ambiguity_results} provides a qualitative validation of increasing the object locational probability examples, which is crucial for the framework described and particularly for interpreting the conditions set. The result illustrates how the system evaluates detections of object locational probability ($o_{ship}$ and $o_{person}$ in this paper) across different UAV camera position, annotating them with an ambiguity level ('High object locational probability' in green, 'Low object locational probability' in red) and the associated locational probability (e.g., $p(o)=0.8$). These annotations are direct outputs of the YOLOv8-world, reflecting the position probability $P(\hat{p}(\mathbf{o}))$ mentioned in Eq. \eqref{eq:prob}).

In subfigures, such as (a)-1, (a)-2, (c)-1, (c)-2, and (d)-1, ships are detected with 'High object locational probability' and correspondingly high locational probability (e.g., $p(o_{ship})$=0.8 in (a)-1; $p(o_{ship})$=0.76 in (a)-2; $p(o_{ship})$=0.64 in (d)-1). These detections would satisfy the condition where the maximum position probability is high, thus being classified as probably approximately correct. For instance, in (a)-1, the clearly visible ship (Score=0.8) is deemed low object locational probability. Conversely, the $p(o_{person})$ detection in the same subplot (Score=0.24) is marked as 'Low object locational probability', implying its confidence score is below the threshold $\sigma_h$, rendering it probably approximately wrong. This demonstrates the system's ability to assign differing object locational probability levels to concurrently detected objects based on their individual detection strengths. Subplot (a)-2 shows a scenario where both the ship (Score=0.76) and a person (Score=0.40) are detected with 'High object locational probability', suggesting that for the $p(o_{person})$ in this instance, a score of 0.40 is considered above the threshold for confident detection.

Challenging scenarios are depicted in subplots (b)-1 and (d)-2. In (b)-1, the overall assessment for the primary object (ship) is 'Low object locational probability' (red text), indicating its detection confidence is likely below $\sigma_h$ due to severe visual interference from waves. The accompanying note "($p(o_{person}) \le$ 0.26)" further highlights the low confidence in any person detection. Similarly, in (d)-2, the ship detection is marked as 'Low object locational probability' in a visually cluttered scene, while a person is detected with a score of 0.55. These instances show the system's critical capability to flag detections with high uncertainty, which is vital for preventing reliance on potentially failed detections in the subsequent multi-UAV path planning.

Subplot (b)-2, labeled "High object locational probability ($p(o)=0.40$)", is particularly illustrative. Despite the challenging visual conditions for ship detection (similar to (b)-1), the system asserts 'High object locational probability' for the 'Person' detection with a score of 0.40. This suggests that even if the primary target (ship) might be highly ambiguous or its confidence low (not explicitly scored for ship here but implied by context), the system can still provide confident assessments for other classes. This aligns with object locational probability correction by indicating that the specific 'Person' detection, with a score of 0.40, is considered "probably approximately correct" for that class.

Collectively, these qualitative examples in Figure~\ref{fig:qualitative_ambiguity_results} support the importance of object locational probability estimation as outlined. The system demonstrates a nuanced understanding by correlating high confidence scores with low object locational probability and low scores with high object locational probability, in line with the relationship between locational ambiguity and correctness described. This capability to differentiate reliable detections from unreliable ones is paramount for the proposed framework's goal of mitigating ambiguity to enhance searching performance and reduce mission failures for multi-UAV operations.

\subsection{Ablation Study of Optimization Gradient}
\begin{figure}[!t]
    \centering
    \includegraphics[width=1\linewidth]{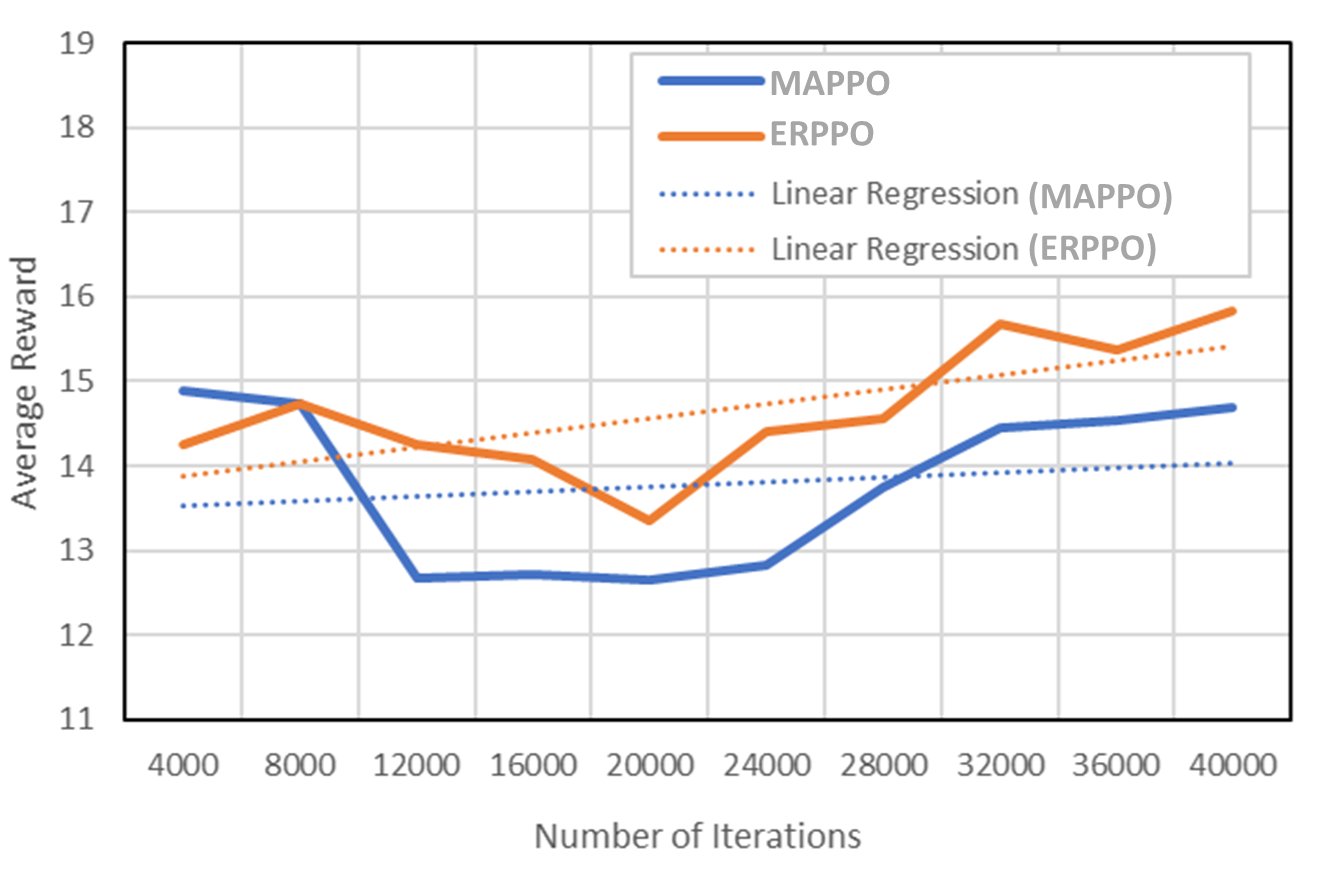}
    \caption{Average reward results of the proposed ERPPO and MAPPO in retraining procedure.}
    \label{fig:am_mappo}
\end{figure}

Figure~\ref{fig:am_mappo} shows the average reward results of the vanilla MAPPO baseline (solid blue) with the proposed ambiguity-mitigation variant, \mbox{ERPPO} (solid orange), over \(4\!\times\!10^{4}\) re-training iterations. Note that this evaluation does not have a scheduled number of iterations for a fair comparison. Both agents start with comparable performance (\(\approx14.8\) average reward at \(4\text{k}\) iterations), yet MAPPO experiences a pronounced degradation between \(8\times 10^3\)–\(2\!\times\!10^{4}\) iterations, bottoming out at \(12.6\).  
ERPPO maintains a markedly higher reward floor (\(13.4\) at \(2\!\times\!10^{4}\)), suggesting that explicit ambiguity modeling misleads observations obtained in the initial exploration phase. From $8\times10^3$ iterations onward, ERPPO outperforms MAPPO at every evaluation point. 
Across the final \(4\!\times\!10^{4}\) iterations, the average reward of ERPPO shows 0.87 higher rewards than basic MAPPO. By following linear regression of ERPPO exhibits a steeper empirical learning slope (0.172) than MAPPO (0.055), equating to a $\times 2.13$ higher rate of performance improvement.

Collectively, these results verify that the proposed ambiguity-mitigation mechanism accelerates policy optimization and yields superior steady-state performance.

\section{Conclusion}
In this paper, we introduced the Entropy Regularization-based Proximal Policy Optimization (ERPPO), a novel approach designed to enhance multi-agent reinforcement learning for adapting entropy in dynamic environments. Our method addresses the challenge of object detection ambiguity arising from non-stationary observations, a common issue that decrease the performance of standard algorithms like MAPPO. We proposed a Distributional Spatiotemporal Ambiguity (DSA) learner to predict object detection ambiguity. By analyzing environmental factors in simulation, the DSA learner provides the RL agent with a crucial inductive bias. We developed a novel entropy-based regularization scheme for PPO. This mechanism dynamically adjusts the learning process by applying a strong L1 regularization in high-ambiguity situations to encourage exploration and a gentle L2 regularization in low-ambiguity states to ensure policy stability. This allows the agents to actively seek out and resolve uncertainty. Experiments conducted in realistic AirSim-based maritime search and rescue scenarios validate the effectiveness of ERPPO. The results demonstrate that the proposed method significantly outperforms existing techniques in both continuous and discrete action spaces. Notably, ERPPO accelerated uncertainty mitigation by up to 2.13 times compared to the baseline and proved superior in suppressing false detections under visually challenging conditions, leading to more stable and higher-performing policies.

\bibliography{aaai2026}


\newpage
\appendix
\section{Supplementary of ERPPO}
\subsection{Proof of Equation~\eqref{eq:ineqa} in the main paper}
\label{proof:ineqa}

As time progresses in the simulation, if an object is continuously influenced by its maritime environment, where waves affect its position, the resulting change in the object's location can be described as a diffusion process over time as follows:
\begin{equation*}
    p(o_k,x,t)=\prod_{i=0}^{t-1}p(o_k,x,i+1|i)
\end{equation*}

As the probability of a specific location outcome satisfies $0 \leq p(o,x,i+1|i)\leq 1$ for all $i$, we can conclude that:
\begin{equation*}
    p(o_k,x,t)\leq p(o_k,x,t+1), \quad \forall t
    \label{eq:ineqa2}
\end{equation*}

\begin{proof}
Probability always follows:
\begin{equation}
    0\leq p(o_k,x,i+1|i) \leq 1, \quad \forall i,\forall o_k, \forall x
    \label{eq:prob}
\end{equation}
Then,
\begin{equation*}
\begin{split}
    p(o_k,x,t+1) &= \prod_{i=0}^{t} p(o_k,x,i+1|i) \\
    &= \prod_{i=0}^{t-1}p(o_k,x,i+1|i) p(o_k,x,t+1|t)\\
    &= p(o_k,x,t) p(o_k,x,t+1|t)\\
    &\geq p(o_k,x,t) \quad \text{by Equation~\eqref{eq:prob}}
\end{split}
\end{equation*}
\end{proof}

\subsection{Detailed loss in terms of reward, value, and policy function in Equation~\eqref{eq:value} and Algorithm~\ref{alg:samegr}}

Multi-Agent Proximal Policy Optimization (MAPPO)~\cite{yu2022surprising} is a variant of the Proximal Policy Optimization (PPO) algorithm~\cite{schulman2017proximal}, specifically tailored for multi-agent reinforcement learning (MARL). MAPPO optimizes cooperative multi-agent settings by employing a centralized critic with decentralized actors. To adapt unlearned ocean environments, our re-training scheme starts from the pretrained policy and value model parameters, $\theta_0$ and $\phi_0$, respectively. For multi-UAVs, $\boldsymbol{u}=\{u^1,...,u^{M_{uav}}\}$, we retrains the RL policy and value model for $K$ steps based on MAPPO~\cite{yu2022surprising} as follows:
\begin{equation}
\begin{split}
    &\phi_{k+1}= \\
    &\arg\min_\phi \frac{1}{|D_k|T}\sum_{\tau\in D_k} \sum^T_{t=0}(V_\phi(Y_t,s_t,\boldsymbol{u}_t^-)-\boldsymbol{R}_t)^2
    \label{eq:mappo11}
\end{split}
\end{equation}
where $o$ presents image observation from an UAV, $s$ shows global states, $\boldsymbol{u}^-$ is all UAVs action set except for an UAV, and $T$ denotes the rollout iterations. $\boldsymbol{R}_t$ is the computed reward as follows:
\begin{equation}
    \boldsymbol{R}_t = \boldsymbol{O}_t(\boldsymbol{Y}(x_t,t,\bar{W}))
\end{equation}
where $\boldsymbol{O}_t$ is the number of ground-truth objects in obtained image $\boldsymbol{Y}(x_t,t,\bar{W})$.
Equation~\eqref{eq:mappo11} shows the critic (or value) model infers the reward given by the observations, global states, and other UAVs' actions sampled from the replay buffer $D_k$.

Here is the updated policy parameter $\theta_{k+1}$ and the original loss $L^{\rm{CLIP}}$ of policy function in MAPPO as follows:
\begin{equation}
    \begin{split}
    \scriptsize
    \theta_{k+1}&=
        \arg\min_\theta L^{\rm{CLIP}}(o,s,u,\boldsymbol{u}^-,\theta_{old},\theta)\\
    &=\arg\min_\theta \Big(\frac{\pi_{\theta}(u|o)}{\pi_{\theta_{old}}(u|o)}A^{\pi_{\theta_{old}}}(o,s,\boldsymbol{u}^-),\\
    & \qquad \qquad \quad  g\big(\epsilon, A^{\pi_{\theta_{old}}}(o,s,\boldsymbol{u}^-)\big) \Big)
    \end{split}
    \label{eq:mappo22}
\end{equation}
where $A^{\pi_{\theta_{old}}}$ is generalized advantage estimation (GAE) and the clip function $g(\epsilon, A)$ equals $(1+\epsilon)A$ if $A\geq0$ or $(1-\epsilon)A$ otherwise. Here, GAE follows:
\begin{equation}
    A_t=\sum^\infty_{l=0} (\gamma \lambda)^l \delta_{t+l}
    \label{eq:GAE}
\end{equation}
where $\delta_t=R_t+\gamma V_\phi(t+1)-V_\phi(t)$. The several hyperparameters are explained in the next section.

\begin{table}
\centering
\resizebox{\linewidth}{!}{%
\begin{tabular}{l|r|c} 
\toprule
Configuration                                              & \multicolumn{1}{l|}{Value} & Unit    \\ 
\hline
The number of UAV agents in the environment.                & 5                          & -       \\
The number of persons to be located by the UAVs.            & 2                         & -       \\
The number of ships.            & 1                         & -       \\
The maximum X and Y boundaries of the search area.          & 500                        & meters  \\
The maximum altitude for the UAVs.                          & 500                         & meters  \\
Simulated positional differences in the DSA                 & 2.5                          & meters  \\
The maximum number of steps before an episode is truncated. & 200                        & steps   \\
The maximum velocity of a UAV in a single step.             & 1                          & m/step  \\
The random value range of a fog weather in Airsim.             & [0.0, 0.3]                          & -  \\
The random value range of a rain weather in Airsim.             & [0.0, 1.0]                          & -  \\
The random value range of wave length in Unreal Engine.             & [521, 6000]                          & -  \\
The random value range of wave amplitude in Unreal Engine.             & [4.0, 80.0]                          & -  \\
\bottomrule
\end{tabular}
}
\vspace{-2mm}
\caption{Configurations of maritime search environment.}
\label{tab:configurations}
\end{table}

\begin{table}
\centering
\scriptsize
\begin{tabular}{l|r|}
\toprule
Hyperparameter       & \multicolumn{1}{l|}{Value}  \\ 
\hline
Gamma ($\gamma$) in GAE in Equation~\eqref{eq:GAE}                   & 0.99                        \\
Lambda ($\lambda$) in GAE in Equation~\eqref{eq:GAE}                   & 0.95                        \\
Learning rate                   & 5e-4                    \\
Epsilon value ($\epsilon$) for clip function in Equation~\eqref{eq:mappo22} & 0.2                         \\
\bottomrule
\end{tabular}
\vspace{-2mm}
\caption{Hyperparameters of ERPPO. All values are re-used for fair comparison of MAPPO.}
\vspace{-4mm}
\label{tab:hype}
\end{table}

\subsection{Configurations and Hyperparameters}
Configurations of the simulated maritime environment are in Table~\ref{tab:configurations}. The number of UAV $N$ is 5. The number of persons to be searched by the UAV is $2$. In our environment, initial position of one ship is shared by multi-UAVs. The search area is defined by $500\times500\times500$. We obtained all images according to the positional differences in $2.5$ meter unit, so the size of DSA map is $200 \times 200 \times 200$. The maximum velocity of UAV is $1$ meter/step. The maximum step of simulation is 200.  The random value range of a fog and rain weather using Airsim is $[0.0, 0.3]$ and $[0.0, 1.0]$, respectively. The maximum fog density is set to $0.3$, as higher values significantly cause a malfunction of object detection. The random value range of wave length and amplitude is $[521, 6000]$ and $[4.0, 80.0]$, respectively.

Hyperparameters are shown in Table~\ref{tab:hype}. The $\gamma$ and $\lambda$ of Equation~\eqref{eq:GAE} is 0.99 and 0.95, respectively. Learning rate is $5e^{-4}$ and epsilon value $\epsilon$ is 0.2 in Equation~\eqref{eq:mappo22}.

\end{document}